\documentclass[sigconf]{acmart}

\usepackage{booktabs} 

\usepackage{caption}
\usepackage{subcaption}

\setcopyright{acmcopyright}

\acmDOI{10.1145/3605098.3635903}

\acmISBN{979-8-4007-0243-3/24/04}

\acmConference[SAC'24]{ACM SAC Conference}{April 8 –April 12, 2024}{Avila, Spain}
\acmYear{2024}
\copyrightyear{2024}

\acmArticle{4}
\acmPrice{15.00}

\begin{document}
\title{Navigating WebAI: Training Agents to Complete
Web Tasks with Large Language Models and
Reinforcement Learning}
\titlenote{Produces the permission block, and
  copyright information}
\subtitle{Extended Abstract}
\subtitlenote{The full version of the author's guide is available as
  \texttt{acmart.pdf} document}
  
\renewcommand{\shorttitle}{Navigating WebAI}

\author{Lucas-Andreï Thil}
\affiliation{%
\institution{Maastricht University}
  \country{the Netherlands}
}
\email{l.thil@student.maastrichtuniversity.nl}

\author{Mirela Popa}
\affiliation{%
  \institution{Maastricht University}
  \country{the Netherlands}
}
\email{mirela.popa@maastrichtuniversity.nl}

\author{Gerasimos Spanakis}
\affiliation{%
  \institution{Maastricht University}
  \country{the Netherlands}
}
\email{jerry.spanakis@maastrichtuniversity.nl}

\renewcommand{\shortauthors}{L. Thil et al.}

\begin{abstract}

Recent advancements in language models have demonstrated remarkable improvements in various natural language processing (NLP) tasks such as web navigation. Supervised learning (SL) approaches have achieved impressive performance while utilizing significantly less training data compared to previous methods. However, these SL-based models fall short when compared to reinforcement learning (RL) approaches, which have shown superior results. In this paper, we propose a novel approach that combines SL and RL techniques over the MiniWoB benchmark to leverage the strengths of both methods. We also address a critical limitation in previous models' understanding of HTML content, revealing a tendency to memorize target elements rather than comprehend the underlying structure. To rectify this, we propose methods to enhance true understanding and present a new baseline of results. Our experiments demonstrate that our approach outperforms previous SL methods on certain tasks using less data and narrows the performance gap with RL models, achieving 43.58\% average accuracy in SL and 36.69\% when combined with a multimodal RL approach. This study sets a new direction for future web navigation and offers insights into the limitations and potential of language modeling for computer tasks.
\end{abstract}

%
%



\maketitle

\section{Introduction}

Neural networks have been used to complete web navigation tasks using multimodal inputs such as Humphrey et al \cite{humphreys2022datadriven} which combined visual and text inputs from a web page. Others focused on pure text and expanded over instruction mapping methods \cite{liu2018reinforcement, pasupat-etal-2018-mapping, he2021actionbert} but were very constrained in their language understanding tasks. The application of language models showed greater capabilities in completing computer tasks, but were all largely constrained to their environment and had poor transfer abilities. Lately, the Cambrian explosion of large language models opened the way for more capable models requiring less data and offering better transfer capabilities as shown by Gur et al. and Kim et al. \cite{gur2023understanding, kim2023language}. Even though they achieved outstanding results over different web navigation benchmarks such as Miniwob++, we highlight in this work that they are in fact more limited than previously claimed. We propose several approaches to rectify some of their issues with memorizing information and highlight the challenges of combining LLMs with a joint multimodal representation of their environment.

Our research primarily focuses on optimizing small-scale models such as grounding them more thoroughly to their environments, in order to benchmark their capabilities before contemplating further scaling. This prioritization aligns with the imperative for fast inference times and adaptability to limited or novel data scenarios.

In this paper, we evaluate the limitations of the Miniwob++ benchmark and the recorded episodes used in previous works in order to present different processing paths to correct their shortcomings. We reproduce the previous results obtained in the literature focusing in the applications of LLMs for web navigation and show that they overfit their benchmark and are unable to recover in slight changes in their environments, hampering the claims that they could understand HTML content. We test our techniques on newly trained models that we enhance by devising hierarchical planning abilities that showcase superior results. Then, we combine them with a multimodal representation using visual inputs that we trained in a supervised manner, and in a reinforcement learning one over multiple phases. We present the abilities of these models to learn such representations, but they suffer from transfer abilities due to the nature of the architectures used.

This paper makes several key contributions. First, we provide insights into the capabilities of agents trained in user interactions to adeptly navigate diverse web interfaces. Second, we introduce a more robust evaluation that exposes the shortcomings of these models due to their tendencies to memorize a lot from their environment and how to overcome them. 
We also set more accurate and grounded results over the Miniwob++ benchmark by taking into account the tendencies of previous work to overfit by applying attacks to the environment. As we showed that previous works tend to learn the distribution of target elements, we provide directions on how to correct it. Lastly, combined with the improvements over our models we deliver a comprehensive analysis of current methods' limitations in order to explore more performant architectures, further contributing to the field toward more capable approaches.


\section{Related Work}
Recent advancements in AI-driven web navigation have led to various approaches to enhance intelligent agents' performance and capabilities. This section highlights key methods and techniques that have shaped this project.

\subsection{OpenAI Universe}
OpenAI's Universe, released in 2016, aimed to develop a general-purpose agent for tasks like video games and web navigation, focusing on Miniwob, a framework with over 80 embedded tasks \cite{openai_universe}\cite{pmlr-v70-shi17a}. Miniwob serves as a reinforcement learning environment, allowing controlled experimentation with web navigation complexities.

\subsection{Neural Network Oriented Approaches}
Several neural-network-oriented models have been devised for web navigation. Notably, Humphrey et al.'s CC-Net combines RL with supervised learning, achieving state-of-the-art results on the Miniwob Benchmark but requiring 2.4 million examples \cite{humphreys2022datadriven}. Liu et al.'s Workflow Guided Exploration, which constrains actions during RL training, has also shown promise \cite{liu2018reinforcement}.

Older works, such as Pasupat et al. \cite{pasupat-etal-2018-mapping}, focused on traditional neural network approaches, while He et al.\cite{he2021actionbert} outlined limitations in multimodal environments. These works have identified challenges and achieved successes but still face generalization issues.

The tradeoff between exploration and exploitation in RL is a common challenge, often requiring extensive data to converge to optimal solutions. While some models, like CC-Net, approach human capabilities, the need for large amounts of data highlights the importance of investigating more efficient models.

\begin{figure}[htbp]
    \centering
    \includegraphics[width=0.40\textwidth]{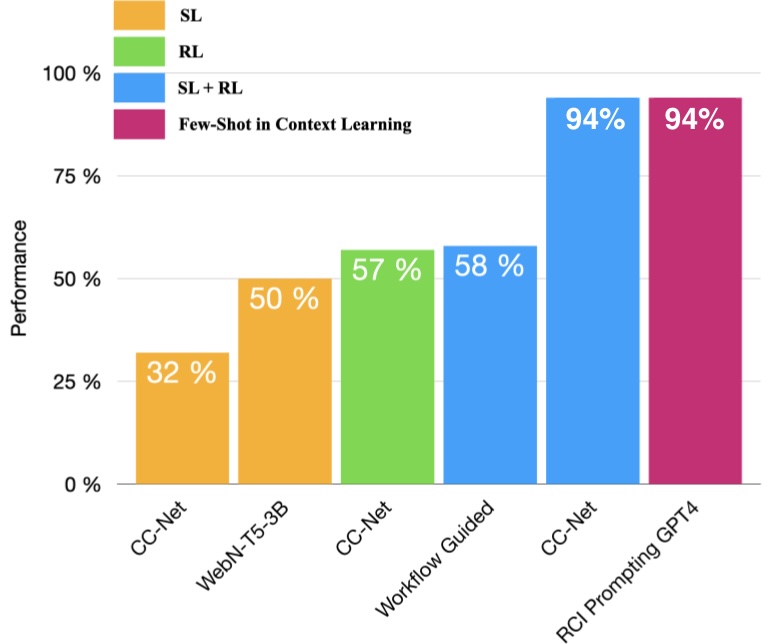}
    \caption{\textbf{Comparison of Existing Models Regarding the Web Navigation Task over the Miniwob Benchmark.} Average comparison over the tasks proposed on the Miniwob benchmark between different training techniques and architectures \cite{humphreys2022datadriven, gur2023understanding, liu2018reinforcement, kim2023language}.}
    \label{fig:comparison_existing_works}
\end{figure}

Figure \ref{fig:comparison_existing_works} illustrates the average accuracy of different models over the Miniwob benchmark, comparing training techniques such as SL, RL, combined approaches, and few-shots prompting examples.

\subsection{Large Language Models}

Early works in web navigation with large language models (LLMs) include Nakano's 'WebGPT' \cite{nakano2022webgpt}, which used GPT-3 \cite{brown2020language} as a browsing assistant but was limited by not using raw web content. Yao et al.'s 'WebShop' \cite{yao2023webshop} created an e-commerce environment for language models but faced restrictions in context window and adaptability.

Attempts to pre-train LLMs on HTML content struggled with benchmarking on classical NLP tasks, limiting their use in web navigation \cite{aghajanyan2021htlm}. However, breakthroughs like Gur et al.'s work on "Understanding HTML with Large Language Models" \cite{gur2023understanding} demonstrated the potential of LLMs in web content comprehension, surpassing previous supervised learning (SL) approaches with significantly less data.

Kim et al. further showcased the potential of larger LLMs like GPT4 \cite{kim2023language, openai2023gpt4} in web navigation tasks through iterative prompting, achieving state-of-the-art baselines in a few-shot manner.

Despite these advancements, challenges remain in developing efficient and adapTable models for web navigation. Existing methods often require extensive training data and may struggle with fast inference times, highlighting the need for smaller, more capable models.

\section{Methods}

Our methodology addresses the limitations in previous large language models (LLMs) for web navigation using the Miniwob benchmark. We begin by describing the environment and dataset, then move to model design in two stages.

In the first stage, we analyze various T5-based models, fine-tuning them with hierarchical planning techniques to overcome identified limitations. In the second stage, we integrate the best-fine-tuned model with a multimodal neural network, using both supervised learning (SL) and reinforcement learning (RL) to enhance performance and adaptability.

We also conduct an ablation study to understand the models' inner workings and identify areas for improvement. Our approach emphasizes assessing performance at a small scale before integrating additional techniques.

\subsection{Miniwob++ Benchmark and Datasets}
\begin{figure}[htbp]
    \centering
    \includegraphics[width=0.45\textwidth]{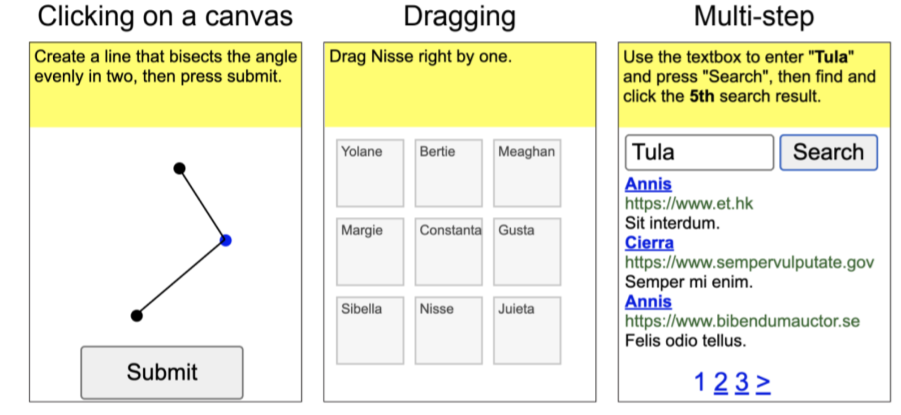}
    \caption{\textbf{Example of Miniwob Episodes.} Each opened episode is timed and alongside it, a discounted reward is computed. These episodes cover a wide range of tasks, and in our case, we select a subset of 40 episodes that are suited to work with language models in the fashion of Gur et al. \cite{gur2023understanding}.}
    \label{fig:miniwob_episode_examples}
\end{figure}

The Miniwob++ benchmark\cite{liu2018reinforcement} offers over a hundred web-based environments, simulating web exploration scenarios \ref{fig:miniwob_episode_examples}. We utilize thirteen thousand human-made demonstrations provided by the Farama Foundation, enabling supervised training. The benchmark's alignment with existing research and compatibility with reinforcement learning (RL) through the gymnasium (previously GYM) environment \cite{brockman2016openai} and techniques makes it suitable for our study.

\begin{figure}[htbp]
    \centering
    \includegraphics[width=0.4\textwidth]{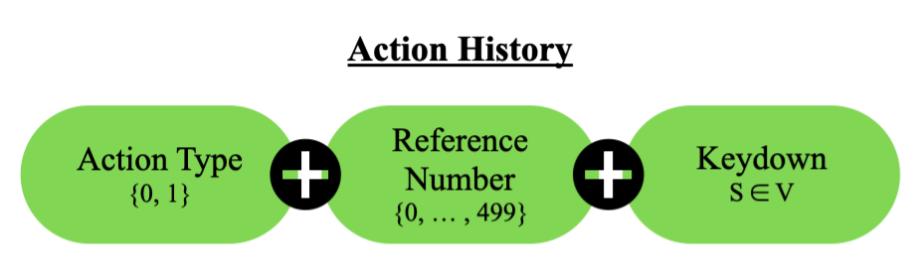}
    \caption{\textbf{Structure of Action History as Proposed by Gur et al. \cite{gur2023understanding}}.}
    \label{fig:action_history1}
\end{figure}

\begin{figure}[htbp]
    \centering
    \includegraphics[width=0.4\textwidth]{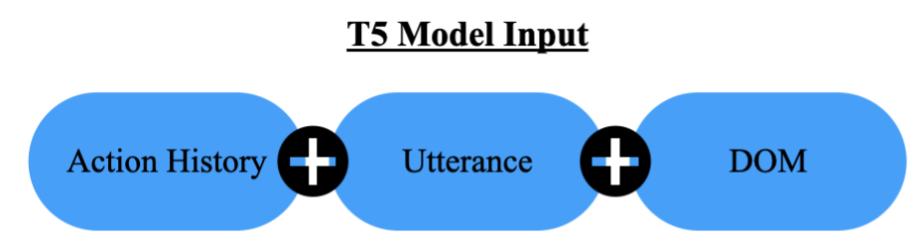}
    \caption{\textbf{Structure of T5 Input in its traditional form as Proposed by Gur et al. \cite{gur2023understanding}}.}
    \label{fig:t5_input1}
\end{figure}

Our datasets include HTML and Document Object Models (DOM) elements parsed as dictionaries, with unique reference numbers ('ref') for identification. We also process mouse interaction data into two actions: 'click' or 'type\_text'. Unlike previous works\cite{humphreys2022datadriven}, we concatenate adjacent typing actions into single actions, see Figures \ref{fig:action_history1} and \ref{fig:t5_input1}. This approach aligns with Gur et al.\cite{gur2023understanding}, using only two separate actions for efficiency and ease of implementation.


Our methodology processes and fine-tunes data from the Miniwob benchmark for web navigation tasks. We create a hierarchical T5 model by identifying sub-tasks within episodes and translating high-level instructions into actionable plans \cite{branavan-etal-2009-reinforcement}. The model is then fine-tuned for planning and action tasks \cite{vogel-jurafsky-2010-learning}, as shown in Figure \ref{fig:T5_example_hierachial}.

\subsubsection{Episode Processing for Action History Extraction}
We process episodes to extract action history, following key steps as seen in Figure \ref{fig:T5_example_hierachial}. The main steps consist of removing duplicate actions, and retaining only the last occurrence. Unnecessary actions, such as clicks on the $<body>$ element, are discarded. Only the last keydown action for each targeted element is retained, with specific cases considered for various interactions. Manual adjustments are made if needed. To address task distribution imbalance, we down-sample 150 episodes in over-represented task suites, as seen in Figure \ref{fig:task_distribution}.

\begin{figure}
\centering
\begin{enumerate}
    \item Infer Model Plan: Devise a plan for the following instruction: 'Click the menu button, and then find and click on the item labeled next' \\
    Subtasks = ['Click menu button', 'find an item labeled next', 'click on the next icon']
    
    \item Loop through the subtasks utterances and infer model: \{Action History, subtask utterance, DOM\} $\rightarrow$ \{Action output, reference, keydown string, boolean final state\}
    
    \item Perform action of the Miniwob environment: Perform action, check if the episode is terminated, and reward
\end{enumerate}
\caption{\textbf{T5 Hierarchical Inference Process.} We first infer the model to devise a navigation plan from the initial utterance, then iterate through the subtask instructions individually to infer the current action at each time step while evaluating the state of the episode by means of the computed reward and terminal state.}\label{fig:T5_example_hierachial}
\end{figure}




\subsubsection{Task Planning Dataset}
We identify and transform various types of sub-actions, such as clicking or selection-based actions. A general function targets each Miniwob episode to produce a list of sub-tasks based on observed interactions. Specific challenges in tasks like flight booking or email forwarding are addressed, with some episodes dropped to ensure clarity. The complexity of some tasks suggests that larger pre-trained language models may perform better.

Here are two examples of the translation between Miniwob utterances and an action sequence for the hierarchical planning task:
\begin{itemize}
\item Example 1:
\begin{itemize}
\item Utterance: {"Departure City":"Philadelphia","Destination City":"Charlotte","Ticket Type":"Return flight","Departure Day":4,"Returning Day":26,"Passengers":2}
\item Action sequence: Select Departure City Philadelphia; Select Destination City Charlotte; Select the Departure Day to 4;
\end{itemize}
\item Example 2:
\begin{itemize}
\item Utterance: Expand the section below and click submit.
\item Action sequence: Expand the section below; click submit;
\end{itemize}
\end{itemize}

\begin{figure*}[htbp]
    \includegraphics[width=0.55\textwidth]{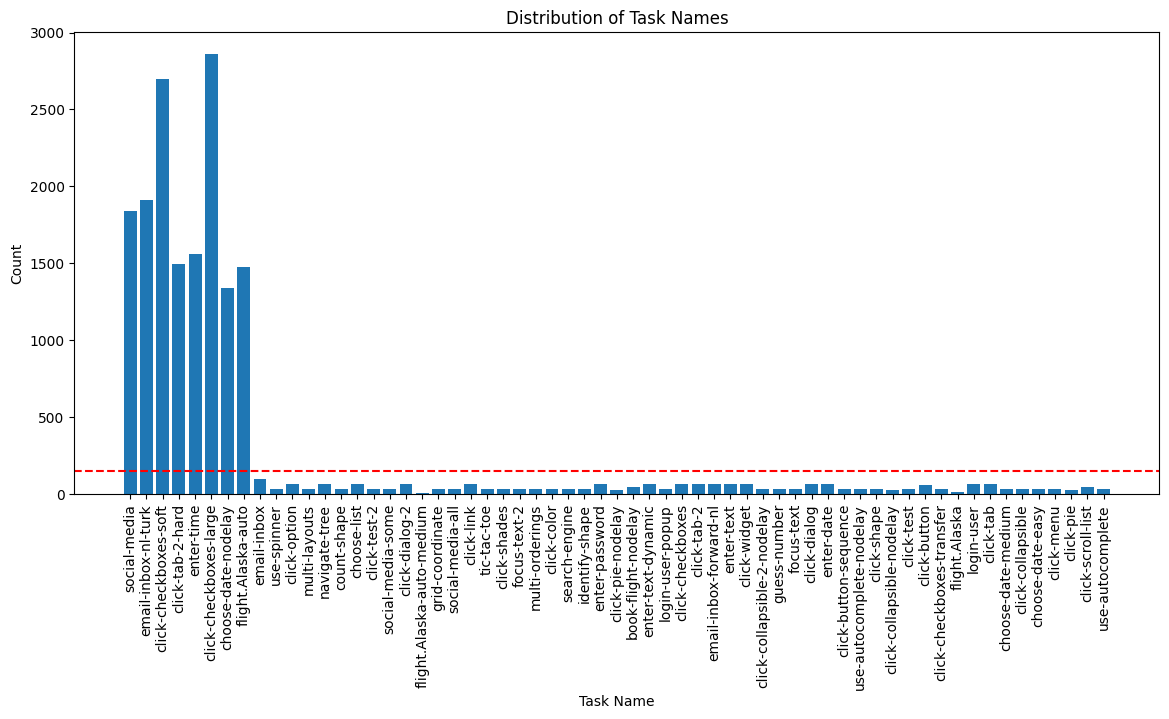} 
    \caption{\textbf{Distribution of task names from the original Miniwob dataset}. We can observe that some tasks are over-represented, therefore we sample a maximum of 150 episodes for the examples exceeding that number. This leads to an average amount of 56.29 of episodes per task in the dataset.}
    \label{fig:task_distribution}
\end{figure*}

The final dataset is composed of over eight thousand action episodes, and in Figure \ref{fig:task_distribution} we describe the number of examples per task. The second dataset regarding task planning instruction contains 10,960 episodes.

\subsection{Models}
In this section, we delve into an array of models designed to tackle web navigation tasks, featuring the WebN-T5 model as our cornerstone, which is based on the T5 architecture and has demonstrated superior performance due to its bi-directional attention encoder \cite{raffel2020exploring, gur2023understanding}. Alongside WebN-T5, we explore variations such as T5-Hierarchy (T5H) fine-tuned for hierarchical tasks, and our hybrid model, CC-NeT5, which combines elements from both T5 and CC-Net by Humphrey et al. \cite{humphreys2022datadriven}. These models have been trained using different strategies like supervised learning (SL) and reinforcement learning (RL), offering a rich comparative landscape for evaluating their effectiveness and limitations in web navigation.


\subsubsection{WebN-T5}
\begin{figure}[htbp]
    \centering
    \includegraphics[width=0.4\textwidth]{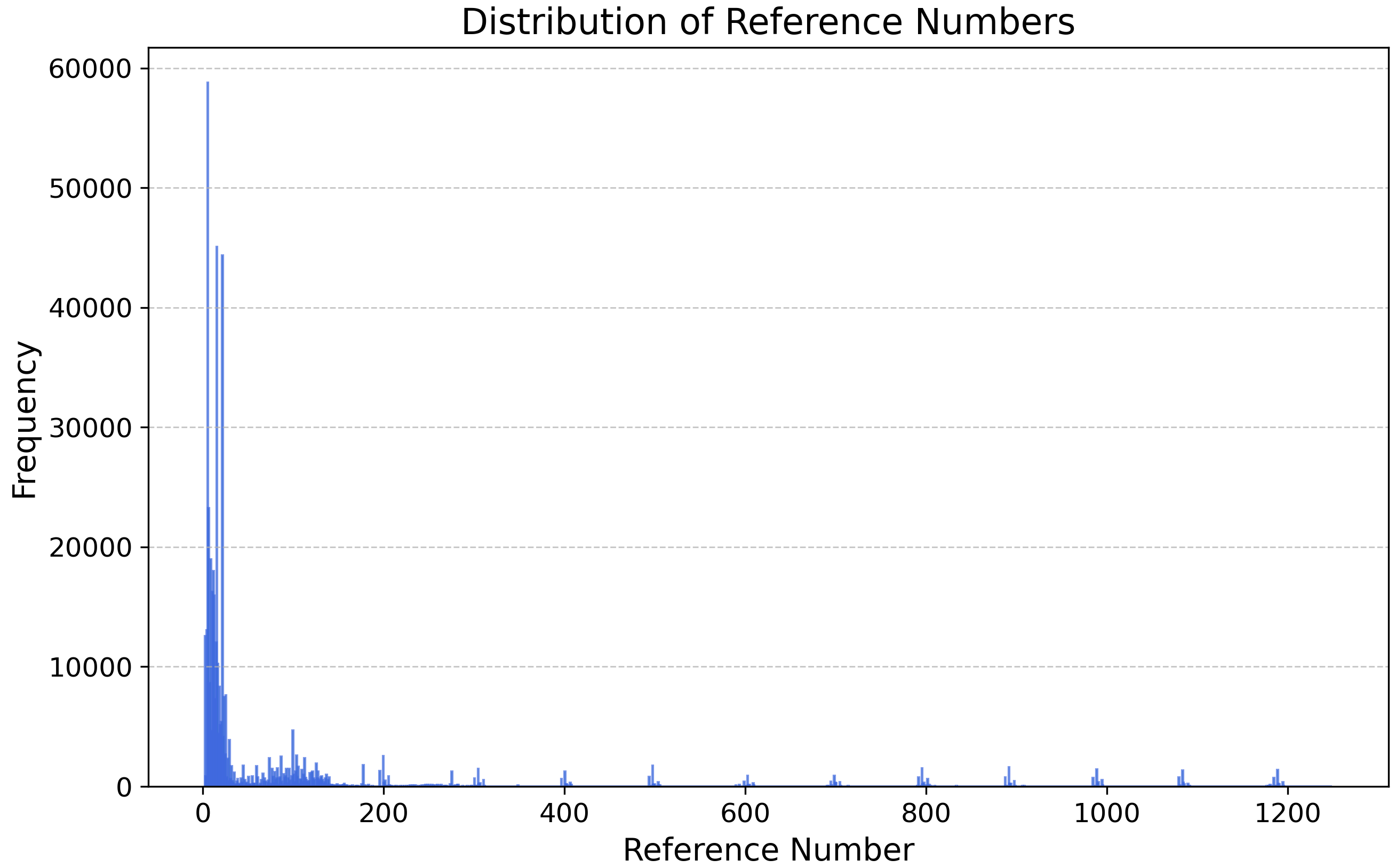} 
    \caption{\textbf{Distribution of Ordered Reference Numbers in the Recorded Actions}. We can observe that the distribution of the target elements is concentrated among several locations, which can be linked to the salient elements in the DOM and displayed on the page. One of our claims is that previous works focused on learning these distributions by overfitting.  }
    \label{fig:ref_distribution_counts}
\end{figure}

We attempt to reproduce the models from 'Understanding Large Language Models' \cite{gur2023understanding} but face challenges with reference numbers and element distribution. As seen in Figure \ref{fig:ref_distribution_counts}, the non-uniform distribution of elements may lead to bias or memorization \cite{carlini2023quantifying} based on their location, or even worse entirely over the distribution of the salient elements of the page \cite{arpit2017closer}. We conduct an experiment with ordered and randomized references to investigate this issue. To mitigate this, we randomize the reference numbers in all episodes, forcing the models to base predictions on element features.

The T5 model was fine-tuned using ROUGE loss metrics \cite{lin-2004-rouge}, which measure the overlap between predicted and reference sequences. However, there are limitations:
\begin{itemize}
\item ROUGE metrics are primarily designed for text summarization or translation, not action sequence prediction.
\item They do not account for the temporal order and structural dependencies in action sequences \cite{schluter-2017-limits}.
\end{itemize}
Thus, while ROUGE offers insights into the quality of generated action sequences, it may not fully capture their correctness and effectiveness.





\subsubsection{WebN-T5 Hierarchical Planning}
We aim to train the agent on lower-level tasks during an episode, addressing observed non-optimal actions in the original WebN-T5 \cite{gur2023understanding}. The model is fine-tuned to divide tasks hierarchically, proposing a multi-step navigation plan by being trained on both datasets containing the task planning and action episodes using a supervised learning approach.

\subsubsection{Multimodal Language Model with Reinforcement Learning}

The best multimodal model used over the Miniwob++ benchmark is CC-Net by Humphrey et al. \cite{humphreys2022datadriven} which achieved human-level performance by using reinforcement learning. Their model used a PPO-based algorithm \cite{schulman2017proximal}, V-MPO \cite{song2019vmpo}, for reinforcement learning (RL) which aims to improve performance without requiring extensive exploration. We used a similar learning approach, except that our model architecture combines a CC-Net-inspired model with a fine-tuned T5 model for the hierarchical planning task detailed earlier. The motivation behind using reinforcement learning, is that many of the recorded examples are not sufficient in covering the variety of the cases proposed. Some environments require further training, and as the exploration space is very large, a method such as V-MPO is better suited to that perspective.

\begin{figure*}[htbp]
    \centering
    \includegraphics[width=0.65\textwidth]{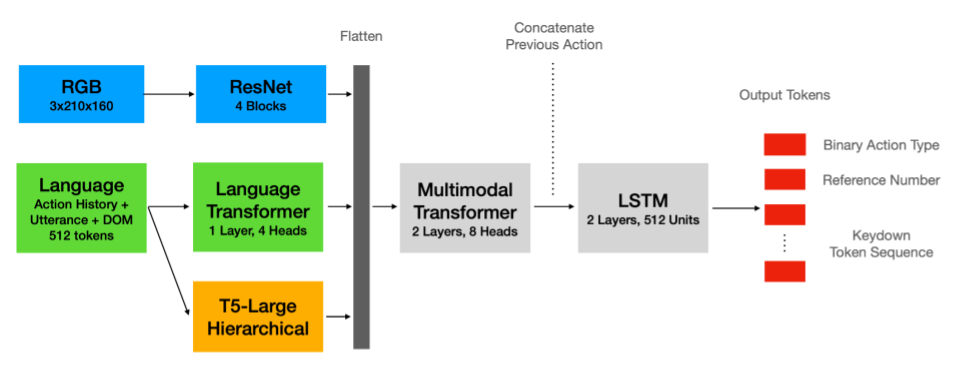}
    \caption{The Combined Architecture of T5-large fine-tuned over the Hierarchical Task, and CC-Net Multimodal Abilities over an RL Approach.}
    \label{fig:ccnet5_architecture}
\end{figure*}

The architecture is derived from CC-Net, using a multimodal approach that includes predictions from the fine-tuned T5 model, a screenshot of the current environment, and language information. The architecture details can be seen in Figure \ref{fig:ccnet5_architecture} and are as follows:
\begin{itemize}
\item Screenshot inputs are processed through four RESNET blocks.
\item Language inputs are embedded and passed through a transformer layer.
\item Outputs are fed into a multimodal block, concatenated with previous actions, and processed through an LSTM block.
\item The final layer includes binary variables for action types and tensors for reference numbers and vocabulary indexes.
\end{itemize}
We adapt the architecture to deal with the vanishing/exploding gradient problem and use one-hot encoding for the experiment.

The design of the CC-NeT5 architecture's loss function was an iterative process, influenced by the nature of its output layer. The final output layer consists of:
\begin{itemize}
\item Action: A binary value represented as a tensor of size 1.
\item Reference number: A tensor of length 500.
\item Keydown text: A tensor of size 8x1591 (8 times our vocabulary size).
\end{itemize}

Initially, a mean-squared-entropy (MSE) loss function was used, but due to sparse encoding and convergence issues, it was updated to a cross-entropy (CE) loss. This change involved predicting reference numbers and keydown text through a softmax function over each section of the output layer. The final loss function efficiently processes this output, with the action type tensor matched with its corresponding boolean value, and the reference number and keydown text token indexes retrieved from the tokenizer.


When using V-MPO, we sample actions from a normalized categorical distribution based on the activation weights of the final layer, where each index represents the token position in our vocabulary. This method allows us to derive a probability distribution for sampling actions during policy inference for exploration. The architecture involves a two-stage process, training the CC-Net-based architecture with V-MPO, while the T5 model remains static. Although the CC-Net part serves as an RL boost to the original T5 model, this approach may have limitations if the original T5 inference is severely flawed.


The model's accuracy is measured in an online environment over the Miniwob benchmark. During training, we alternate between offline (SL) and online (RL) phases \cite{agarwal2020optimistic}, using recorded Miniwob episodes for the first offline phase and successful episodes from online RL phases for subsequent offline ones \cite{schrittwieser2021online}.

We propose a new preset in Gymnasium's RL environment, reducing the action space to two types and adjusting the observation space. The time limit for episodes is increased to thirty seconds, and a discounted reward is computed to train the models in an RL manner.

\section{Results}

This section analyzes the outcomes of various trained models, including ablation studies, and compares them with existing models. The models are benchmarked on the Miniwob benchmark, focusing on click and typing actions, and the accuracy is measured over a hundred episodes.

\subsection{Model Performances}
Our experiments reveal that the T5-large model, fine-tuned on hierarchical tasks, achieves superior performance with an average accuracy of 43.58\%. This outperforms the T5-base model, which reaches an average accuracy of 39.78\%. Interestingly, a hybrid approach combining T5-large with a CC-Net-inspired architecture yields an accuracy of 36.39\% in its BC phase. However, this drops to 33.86\% after the RL phase, a phenomenon we discuss in subsection \ref{sssec:rrlp} due to a covariate shift towards the T5 model. A comparison of the different models' performance metrics are presented in Figure \ref{fig:baseline_results_t5}, including the ablation study over their inputs.

\begin{figure*}[htb]
    \centering
    \begin{minipage}{0.6\textwidth}
        \centering 
    
        \begin{minipage}[b]{0.30\textwidth}
            \centering
            \includegraphics[width=0.8\textwidth]{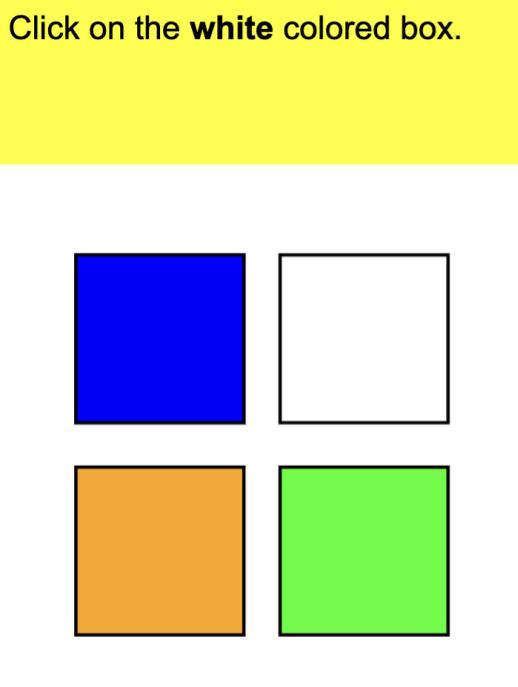}
            \caption{\textbf{Episode example of choosing one out of four colors displayed from the original utterance}. }
            \label{fig:choose_color_episode}
        \end{minipage}
        \hfill
        \begin{minipage}[b]{0.30\textwidth}
            \centering
            \includegraphics[width=0.8\textwidth]{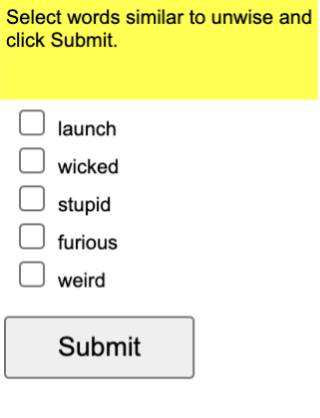}
            \caption{\textbf{Episode example of checking boxes of words related to the one given in the original utterance.}}
            \label{fig:click-checkboxes-soft}
        \end{minipage}
        \hfill
        \begin{minipage}[b]{0.30\textwidth}
            \centering
            \includegraphics[width=0.8\textwidth]{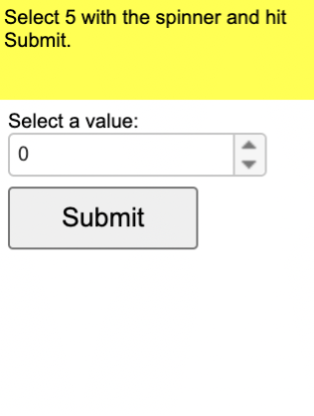}
            \caption{\textbf{Increment the spinner to a number of the desired value and click submit episode example.}}
            \label{fig:use-spinner}
        \end{minipage}
    \end{minipage}
\end{figure*}

\subsubsection{T5-Model Performance}
Fine-tuned T5 models show variable performance. T5-large scores 43.49\% accuracy on hierarchical tasks, while its ablated version slightly edges it with 43.58\%. In contrast, T5-base lags with 39.77\% accuracy as seen in Figures \ref{fig:bc_results_t5_base_hierarchical} and \ref{bc_results_t5_large_hierarchical}. The performance advantage in tasks like 'click-checkboxes-soft' indicates that larger models capitalize better on their pre-trained linguistic skills.

\begin{figure*}[ht]
    \centering
    \includegraphics[width=0.75\textwidth]{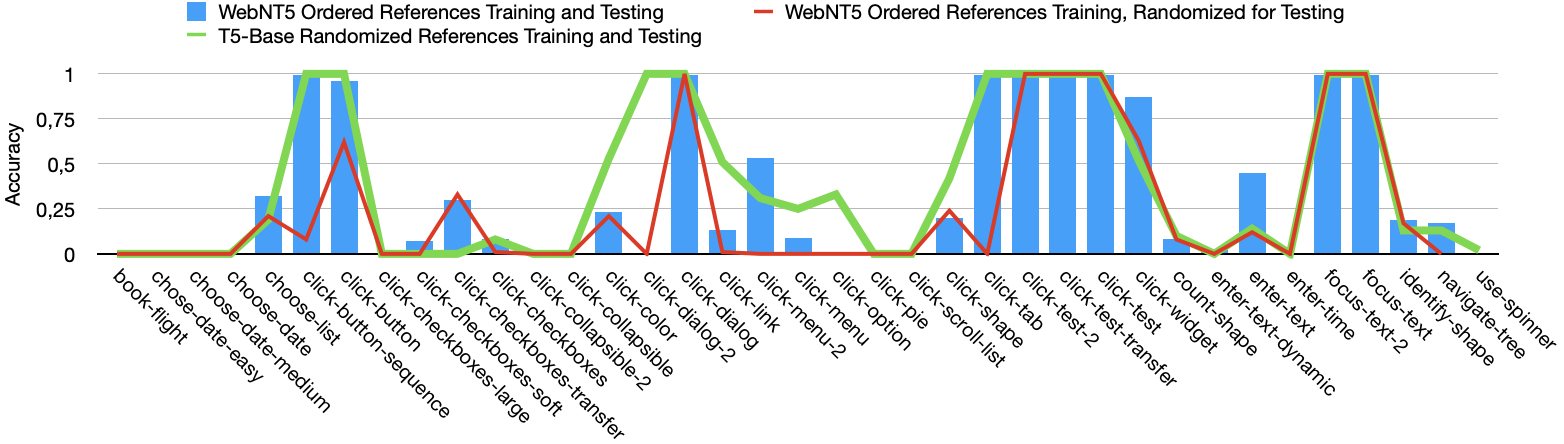}
    \caption{Comparison and effects of fine-tuning with ordered references on a randomized reference test set, and when directly fine-tuned with randomized references.}
    \label{fig:comparison}
\end{figure*}

Our replication of WebN-T5 by Gur et al. \cite{gur2023understanding} found that model performance hinges on reference ordering. Training with randomized references improved its performance, validating our randomization process, as depicted in Figure \ref{fig:comparison}.

\subsubsection{Evaluation of Original Papers}
Our findings upon reproducing Gur et al.'s work \cite{gur2023understanding} reveal memorization tendencies rather than genuine task understanding. Randomizing references resulted in performance drops, questioning the original claims but reaffirming the importance of data randomization in model training.

\begin{figure}[htbp]
    \centering
    \includegraphics[width=0.35\textwidth]{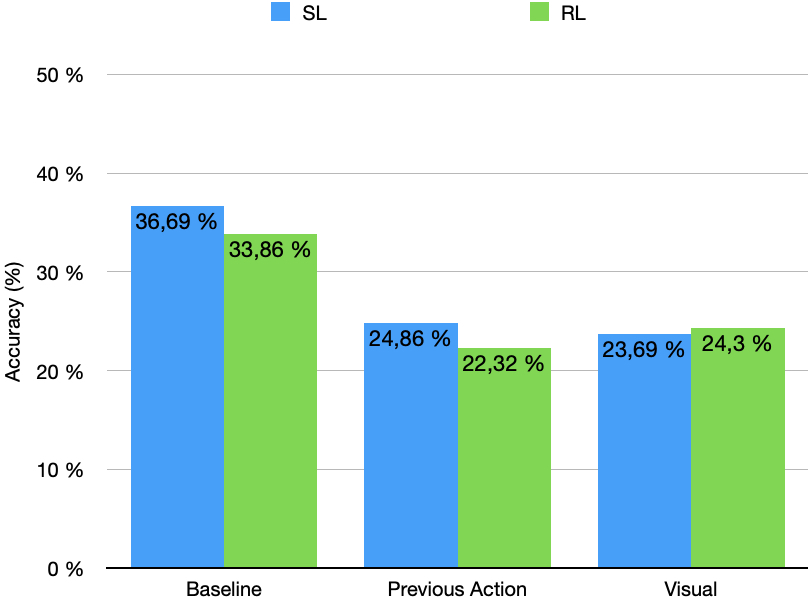}
    \caption{Ablation study of CC-Net5 after its initial SL phase and final RL one.}
    \label{fig:ablation_ccnet5}
\end{figure}
\subsection{History versus No History}
An ablation study on action history reveals interesting insights. The T5-base model's performance drops by nearly two percentage points when action history is removed. In contrast, models fine-tuned on hierarchical tasks, such as T5-large, show almost no performance change, as detailed in Table \ref{tab:language_model_results} and Figures \ref{fig:bc_results_t5_base_hierarchical} and \ref{bc_results_t5_large_hierarchical}. This suggests that the hierarchical nature of the task allows the model to plan its actions more effectively, making it less reliant on action history. Furthermore, the randomization of references seems to make the model less dependent on previous sequences, providing another layer of resilience to the removal of action history.

\subsection{Hierarchical Planning Improvements}

Fine-tuning a model over the hierarchical task achieved higher results than the compared WebN-based models. The following T5 sizes have been fine-tuned which are T5-base and T5-large, achieving respectively  42.1\% and 43.49\% accuracy shown in Table \ref{tab:language_model_results}. The original WebN achieved 46.4\% accuracy over WebT5-large, and the reproduced T5-base size achieved 38.1\% which shows that hierarchical planning is an important component of these models.

The ablation study over the action history outlined a 14\% decrease in performance over the T5-base model for hierarchical planning and 0.2\% over the T5-large. This is much less than the reported 6.4\% reported by the original WebN-T5-large and WebN-T5-3B models showing that hierarchical planning is less sensitive to the action history as it tries to solve the episode by following an original plan step by step.

\begin{figure*}[ht]
    \centering
    \begin{subfigure}[b]{0.75\textwidth}
        \includegraphics[width=\textwidth]{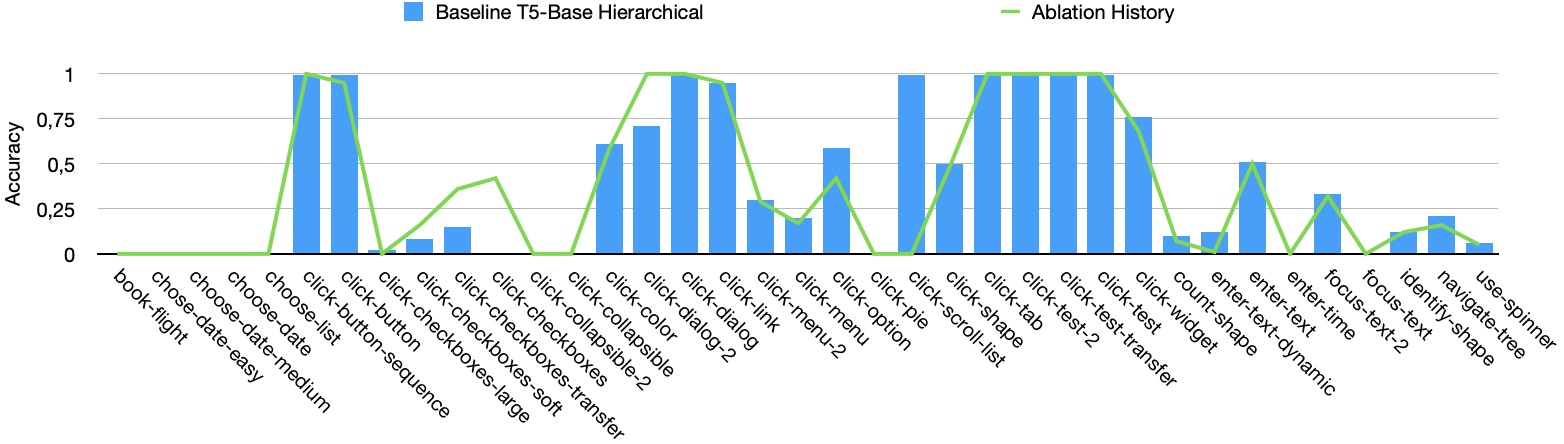}
        \caption{Comparative results of T5-base fine-tuned over the navigation task over the Miniwob++ benchmark with the ablation of its action history.}
        \label{fig:bc_results_t5_base_hierarchical}
    \end{subfigure}
    \hfill  
    \begin{subfigure}[b]{0.75\textwidth}
        \includegraphics[width=\textwidth]{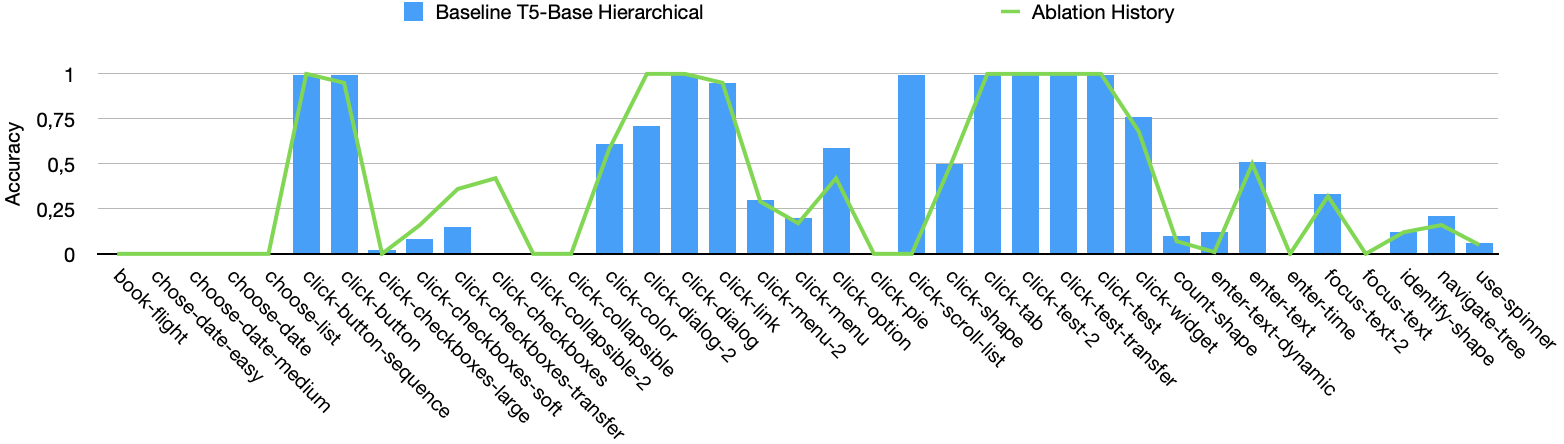}
        \caption{Comparative results of T5-large fine-tuned over the navigation task over the Miniwob++ benchmark with the ablation of its action history.}
        \label{bc_results_t5_large_hierarchical}
    \end{subfigure}
    \caption{Comparative results of the different T5-only models over the navigation task with an ablation study of their inputs over the Miniwob++ benchmark.}

    \label{fig:baseline_results_t5}
\end{figure*}

Nonetheless, we do observe significant rates of failures in complicated episodes, or when the environment changes as the agent follows the original plan as seen in detail in Figures \ref{fig:bc_results_t5_base_hierarchical} and \ref{bc_results_t5_large_hierarchical}

\begin{figure*}[ht]
    \centering
    \begin{subfigure}[b]{0.75\textwidth}
        \includegraphics[width=\textwidth]{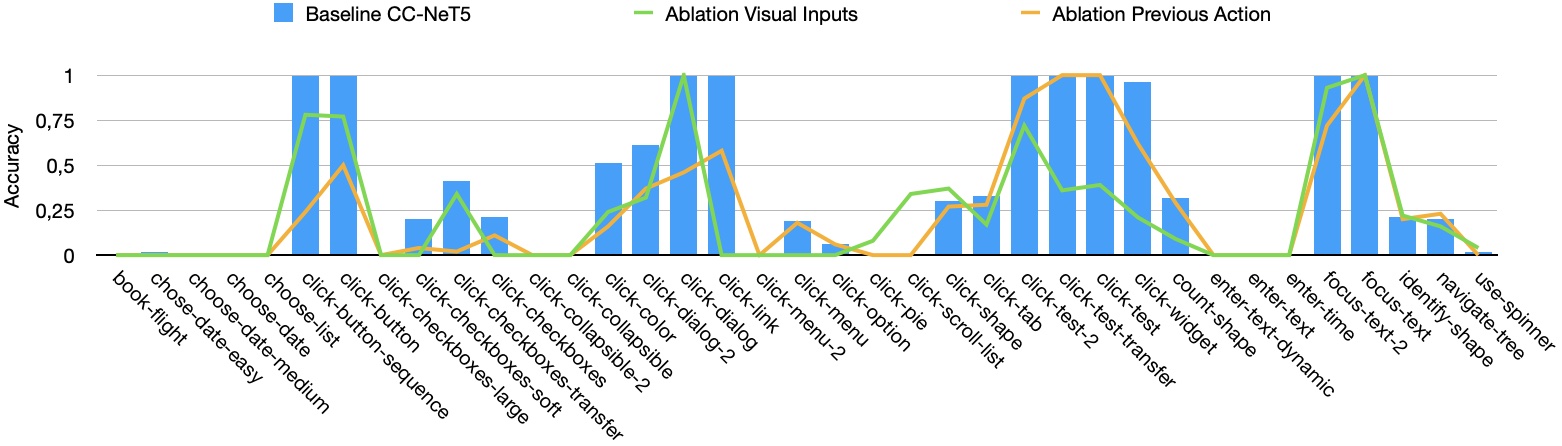}
        \caption{Comparative results of CC-NeT5 after the initial BC phase over the Miniwob++ benchmark with the ablation of its visual and previous action inputs.}
        \label{bc_results_ccnet5}
    \end{subfigure}
    \hfill
    \begin{subfigure}[b]{0.75\textwidth}
        \includegraphics[width=\textwidth]{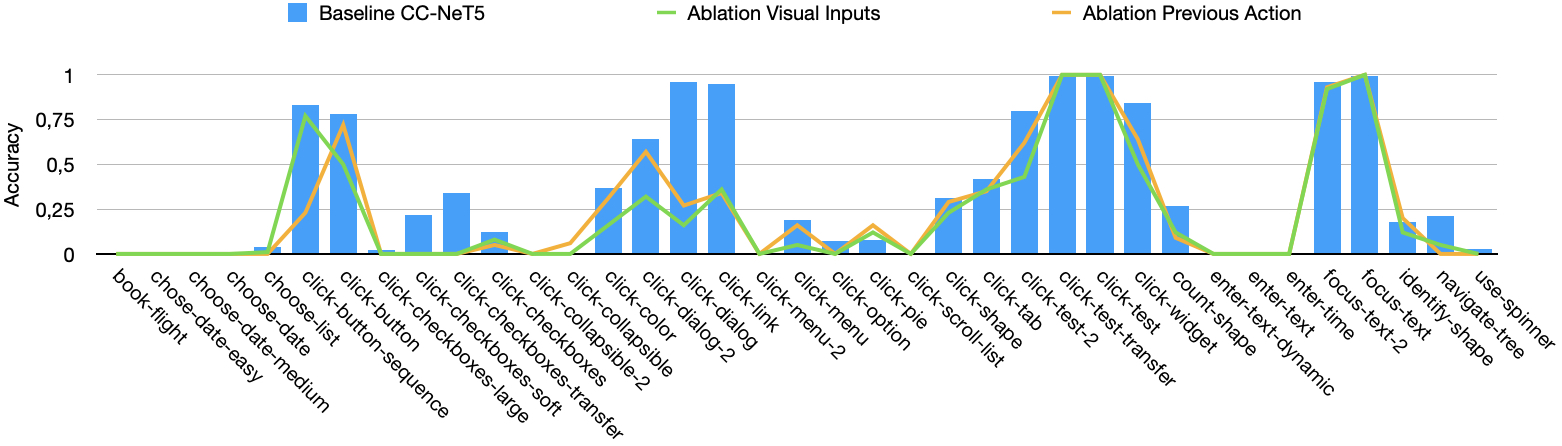}
        \caption{Comparative results of CC-NeT5 after the final RL phase over the Miniwob++ benchmark with the ablation of its visual and previous action inputs.}
        \label{rl_results_ccnet5}
    \end{subfigure}
        
    \caption{Comparative results of the CC-NeT5 architecture over the navigation task with an ablation study of their inputs over the Miniwob++ benchmark for both supervised and reinforcement learning.}
    \label{fig:baseline_results_ccnet5}
\end{figure*}

\subsection{Performance of Combining T5 and CC-Net}
The combination of T5 and CC-Net was explored in two phases: supervised learning (SL) and reinforcement learning (RL), with ablation studies conducted in both.

\subsubsection{Results of Supervised Learning Phase}
The initial SL phase achieved 36.69\% accuracy, lower than the best T5-large model observed in Table \ref{tab:language_model_results}. Ablation studies revealed that the model was heavily dependent on the T5-large model, with complete failure when T5's output was removed. Interestingly, the removal of visual inputs sometimes improved performance as seen in Figure \ref{bc_results_ccnet5}, indicating a complex interaction between modalities. Overall, the results showed that the model learned a multimodal representation but was mainly reliant on the T5-large model.

\subsubsection{Results of Reinforcement Learning Phase}\label{sssec:rrlp}
The RL phase resulted in a further drop in accuracy to 33.86\% observed in Table \ref{tab:language_model_results}. Ablation studies confirmed the model's continued dependence on the T5-large model as seen in Figures \ref{fig:ablation_ccnet5} for average accuracy and \ref{rl_results_ccnet5} for specific tasks. Several factors may have contributed to this decline, including the model's inherent complexity, sensitivity to hyperparameter tuning presented in Table \ref{tab:vmpo_parameters}, the nature of the RL environment, and possible covariate shift between SL and RL phases. These challenges highlight the intricacies of RL training and the need for careful analysis and tuning.

\subsection{Benchmarking Results}
The model was benchmarked on a subset of Miniwob tasks, focusing on 40 out of 80 available tasks. The results showed competitive performance with previous supervised learning methods while using less training data. 
A key finding was the tendency of previous models to memorize rather than understand the distribution of target elements. This work's benchmarking results, though sometimes lower, offer a more robust and reliable assessment of performance.


\section{Discussion and Limitations}

Models trained solely on Miniwob are proficient but lack transferability to diverse web tasks. Pretrained models like T5 have limitations in input context and are prone to overfitting. Our approach in Miniwob++ establishes a more realistic baseline, albeit sometimes lower than prior works \cite{ziegler2020finetuning, NIPS2017_d5e2c0ad}, but we showed it is more grounded to its environment.

Performance evaluation needs to be multifaceted, incorporating metrics beyond task accuracy, such as generalization abilities. Models remain data-intensive and susceptible to memorization; alternative approaches like RLHF should be explored \cite{ziegler2020finetuning, NIPS2017_d5e2c0ad}. 
Despite their limitations, large pre-trained models still perform best, but intermediate-sized models remain under-explored. Small models excel in task-constrained settings but struggle with complexity, while large models offer better generalization but are often overqualified for tasks they solve \cite{hoffmann2022training, lecun_doom}. The integration of multimodal approaches, like ours with T5, reveals tokenization discrepancies that need optimization \cite{hoffmann2022training, lecun_doom}, and training improvements may include randomizing DOM elements and using ablated inputs to focus on content rather than pattern memorization.

Web navigation automation involves critical ethical and legal aspects that demand attention. Ensuring user privacy is paramount, requiring secure data handling and compliance with regulations like GDPR \cite{GDPR2016a}. The growing capability of large language models to mimic humans raises ethical concerns about impersonation \cite{baudrillard}, outpacing traditional identifiers like the Turing test \cite{ayesh2019turing}. Finally, clear accountability must be established to navigate complex liability issues, including potential violations of copyright laws and data protection policies. These challenges highlight the need for robust regulations and well-defined licensing agreements.

\section{Conclusion}



Automating web navigation tasks offers many benefits but also presents significant challenges.

Our behavioral cloning and hierarchical planning models achieved a top accuracy of 43.58\% on the Miniwob++ benchmark, setting a more grounded performance baseline by mitigating overfitting tendencies commonly seen in large language models (LLMs). While the fine-tuned multimodal model excelled in behavioral cloning, its reinforcement learning phase was hindered by covariate shift due to architectural limitations. We highlight the need for further exploration in multimodal architecture design and the potential of fine-tuned LLMs. 

 This work 
 emphasizes the importance of understanding the limitations of LLMs over web navigation tasks, exploring optimal architectures, and considering ethical implications. While large models excel in few-shot scenarios, smaller models are efficient for known environments. The exploration of intermediate sizes and multimodal models offers promising opportunities. Pre-processing techniques can alleviate some issues, but further efforts are needed for more capable models. Ethical considerations, including misuse, impersonation, and accountability, must be addressed to ensure the safety and integrity of the web. This work contributes by highlighting these challenges and proposing techniques to overcome them, paving the way for future advancements in the field.


\section{Appendix}

\begin{table}[h]\footnotesize
\centering
\caption{\textbf{Average Accuracy of the Different Models, Fine-Tuned T5 and Combined with CC-Net.}}
\begin{tabular}{|c|c|}
\hline
\textbf{Model Name} & \textbf{Average Accuracy} \\ \hline
T5-Base Hierarchical & 39.77\% \\ \hline
T5-Base Hierarchical No-History & 38.02\% \\ \hline
T5-Large Hierarchical & 43.49\% \\ \hline
T5-Large Hierarchical No-History & 43.58\% \\ \hline
WebNT5-Base & 37.94\% \\ \hline
WebNT5-Base No-History & 35.99\% \\ \hline
WebNT5-Base Ordered Refs & 35.25\% \\ \hline
WebNT5-Base Randomized Refs Test & 24.19\% \\ \hline
CC-NeT5 Hierarchical (SL) & 36.69\% \\ \hline
CC-NeT5 Hierarchical (SL+RL) & 33.86\% \\ \hline
\end{tabular}
\label{tab:language_model_results}
\end{table}

\begin{table}[ht]
\centering
\caption{\textbf{Hyper-parameters used in V-MPO during the RL 'online' phase of our training, inspired by CC-Net training parameters.}}
\begin{tabular}{|l|l|}
\hline
\textbf{Parameter} & \textbf{Value} \\
\hline
Optimizer & Adam \cite{kingma2017adam} \\
Learning rate & 1e{-4} \\
Adam b1 parameter & 0.9 \\
Adam b2 parameter & 0.999 \\
Weight decay (biases excluded) & 1e{-1} \\
VMPO $\alpha$ & 0.1 \\
VMPO $\eta$ & 0.2 \\
Agent discount $\gamma$ & 0.9 \\
Batch size SL & 120 \\
Trajectory unroll length & 64 \\
Target-network update period $T$ & 5 \\
Maximum number of steps per episode & 10
\\
\hline
\end{tabular}
\label{tab:vmpo_parameters}

\end{table}

\bibliographystyle{ACM-Reference-Format}
\bibliography{sample-sigconf} 

\end{document}